\def\year{2022}\relax
\documentclass[letterpaper]{article} 
\usepackage{aaai22}  
\usepackage{times}  
\usepackage{helvet}  
\usepackage{courier}  
\usepackage[hyphens]{url}  
\usepackage{graphicx} 
\usepackage{natbib}  
\usepackage{caption} 
\DeclareCaptionStyle{ruled}{labelfont=normalfont,labelsep=colon,strut=off} 
\usepackage{algorithm}
\usepackage[noend]{algorithmic}
\usepackage{booktabs}
\usepackage[super]{nth}

\usepackage{newfloat}
\usepackage{listings}
\floatstyle{ruled}
\newfloat{listing}{tb}{lst}{}
\floatname{listing}{Listing}
\usepackage{soul}
\usepackage{xcolor}
\usepackage{amsmath}
\usepackage{amsfonts}

\newcommand{\eat}[1]{} 

\title{Top-Down Deep Clustering with Multi-generator GANs}
\author{
Daniel de Mello \textsuperscript{\rm 1}, 
Renato Assun\c{c}\~{a}o \textsuperscript{\rm 2, 1}, 
Fabricio Murai \textsuperscript{\rm 1}}
\affiliations {
    \textsuperscript{\rm 1} Department of Computer Science, Universidade Federal de Minas Gerais\\
    \textsuperscript{\rm 2} ESRI Inc. \\
    dani-dmello@hotmail.com, rassuncao@esri.com, murai@dcc.ufmg.br
}
\usepackage{bibentry}

\begin{document}

\setlength{\abovedisplayskip}{3pt}
\setlength{\belowdisplayskip}{3pt}
\setlength{\textfloatsep}{5pt}

\maketitle

\begin{abstract}

Deep clustering (DC) leverages the representation power of deep architectures to learn embedding spaces that are optimal for cluster analysis. This approach filters out low-level information irrelevant for clustering and has proven remarkably successful for high dimensional data spaces. Some DC methods employ Generative Adversarial Networks (GANs), motivated by the powerful latent representations these models are able to learn implicitly. In this work, we propose HC-MGAN, a new technique based on GANs with multiple generators (MGANs), which have not been explored for clustering. Our method is inspired by the observation that each generator of a MGAN tends to generate data that correlates with a sub-region of the real data distribution. We use this clustered generation to train a classifier for inferring from which generator a given image came from, thus providing a semantically meaningful clustering for the real distribution. Additionally, we design our method so that it is performed in a top-down hierarchical clustering tree, thus proposing the first hierarchical DC method, to the best of our knowledge. We conduct several experiments to evaluate the proposed method against recent DC methods, obtaining competitive results. Last, we perform an exploratory analysis of the hierarchical clustering tree that highlights how accurately it organizes the data in a hierarchy of semantically coherent patterns.

\end{abstract}

\eat{
\section{LEGENDA DE MARCAÇÕES NO TEXTO}
\begin{itemize}
    \item{Texto sem marcação: texto retirado da dissertação e já revisado}
    \item {\color{black} Azul: texto novo \textit{ou} que estava na dissertação mas não chegou a ser revisado}
    \item {\color{red} Vermelho: comentário}
     
\end{itemize}
}

\section{Introduction}
\eat{
{\color{red} (Daniel) Obsv: O texto sem marcação nesta introdução foi revisado e retirado da dissertação, porém, ele foi revisado somente quando estava em uma versão em português.}
}

Cluster analysis is a fundamental problem in unsupervised learning, with a wide range of applications, especially in computer vision \cite{cluster_app_1, cluster_app_2, cluster_app_3, cluster_app_4}. Its goal is to assign similar points of the data space to the same cluster, while ensuring that dissimilar points are placed in different clusters. One of the main challenges in this approach is to quantify the similarity between objects. For low-dimensional data spaces, similarity might be straightforwardly defined as the minimization of some geometric distance (\textit{e.g.} euclidean distance, squared euclidean distance, Manhattan distance). On the other hand, choosing the distance metric becomes unfeasible for high dimensional data distributions. Images are a clear example of this problem, since any distance metric based on raw pixel spaces is subject to all sorts of low-level noisy disturbances irrelevant for determining the similarity and suffer from a lack of translation or rotation invariance. This motivates the need for some dimensionality reduction technique, by which the fundamental relationships between objects projected onto the resulting embedding space become more consistent with geometrical distances.

In recent years deep clustering techniques have spearheaded the dimensionality reduction approach to clustering by employing highly non-linear latent representations learned by deep learning models \cite{deep_learning_1}. Considering the unsupervised nature of cluster analysis, the models that naturally arise as candidates for deep clustering are unsupervised deep generative models, since these must learn highly abstract representations of the data as a requirement for realistic and diverse generated samples. One of such models are the  Generative Adversarial Networks (GAN) \cite{gan}, whose extremely realistic results in image generation, semantic interpolation and interpretability in the latent space, are evidence of their capacity of learning a powerful latent representation that captures the essential components of the data distribution. Nonetheless, very few works have proposed GAN architectures designed for clustering. Some of these works are  \cite{clustergan,infogan}, where the authors showed that, by manipulating the generator's architecture in a specific way, it is possible to control the class of the training data to which a generated sample belongs, even when classes labels are not available during the training.  

Some recent works employed a GAN architecture with multiple generators \cite{madgan, mgan, stackelberg} to achieve greater diversity in image generation, as well as an alternative way of stabilizing the training. In these works, the authors have observed that each generator tends to specialize in generating examples belonging to a specific class of the dataset. This suggests the organic emergence of clusters in the generators' representation of the data and it is one of the main motivations of our work. The rationale is that clustering would be possible by employing a classifier in charge of distinguishing between the generators, and this classifier could later be applied to the real dataset in order to classify real examples without the use of labels. Additionally, the problem of setting the number of generators to be used for representing the classes of the training set is not addressed in these works, which could be an issue in a real clustering task, where the number of clusters is assumed to be unknown.

\eat{
{\color{red} (Daniel) Eu gostaria de tentar depois melhorar a justificativa no próximo parágrafo para o emprego do método hierárquico. }
}

In this work, we propose Hierarchical Clustering MGAN (HC-MGAN), a method that leverages the multi-generator GAN for the clustering task. We employ multiple generators, each of them specializing in representing a particular cluster of the training distribution. This should lead to a  stronger representation capacity and with more meaningful clusters than what a single generator covering multiple clusters can provide. MGANs have not been used in the previous works exploring GANs for clustering. 
Additionally, we design our method so that it performs the clustering of the training data in a top-down hierarchical way, creating new generators as divisions of subsequent clusters become necessary, \textit{i.e.}, it permits the user to control different clustering granularity levels according to the task at hand.  
The main contributions of this work are as follows:
\begin{itemize}
\item We propose HC-MGAN, a novel deep clustering method that employs GANs with multiple generators.
\item We design HC-MGAN as a top-down hierarchical clustering tree introducing the first hierarchical deep clustering algorithm. Hierarchical clustering allows the user to control different levels of cluster granularity and favors interpretability by taxonomically organizing the clusters.
\item We conduct experiments with three image datasets used in other deep clustering works as benchmarks, namely, MNIST, Fashion MNIST and Stanford Online Products. We obtain competitive results against recent deep clustering methods, all of which are horizontal and therefore lack the advantages of the hierarchical approach.
\item We explore the clustering pattern obtained throughout the tree, displaying how HC-MGAN is able to organize the data in a semantically coherent hierarchy of clusters.
\end{itemize}

\section{Related Work}\label{ref}

\eat{
{\color{red} (Daniel) Vou falar mais a fundo das MGANs, com equações, na metodologia. Os papers de MGANs já foram mencionados na introdução, acho que não há necessidade de incluir no Related works, que já está bem cheio. } Renato: OK, por mim, já pode apagar tirar este comentario.
}

{\color{black}We review some recent work in topics related to Deep Clustering and GANs. 
Autoencoders (AEs) have been the most prevalent choice in the deep clustering literature \cite{dec, dcn, benchmark:DualAE, benchmark:NCSC}, where the clustering objective is usually optimized on the feature representation $\mathbf{Z}$ resultant of the mapping $E:\mathcal{X}\rightarrow\mathcal{Z}$ learned by the encoder component in AEs for a data space $\mathcal{X}$ and a feature space $\mathcal{Z}$. Deep Embedded Clustering (DEC) \cite{dec} pioneered this approach, obtaining state-of-the art clustering results. It works by pretraining an AE with a standard reconstruction loss function, and then optimizing it with a regularizer based on the clustering assignments modeled by a target student-t distribution, having the cluster centers iteratively updated. DEC's results were surpassed by \cite{dcn}, which converted DEC's objective function into an alternated joint optimization with K-Means loss, thus obtaining a clusterization more suited for K-Means.}

The use of GANs for clustering tasks has been influenced by InfoGAN \cite{infogan}, a type of GAN whose latent variable consists of, besides the usual multidimensional variable $z$, a set $c$ of one-dimensional variables $c_1, c_2 ... c_N$ that are expected to unsupervisedly capture semantic information in a disentangled manner (\textit{i.e.}, with each variable encoding isolated interpretable features of the real data). For obtaining this, the authors of InfoGAN proposed an additional term in the generator's loss function that maximized the mutual information $I(c;G(z,c))$ between a generated image $G(z,c)$ and the latent variable $c$ that originated it. The variables $c$ could be chosen to represent both categorical and continuous features. 
ClusterGAN \cite{clustergan} is an architecture appropriate for clustering tasks based on InfoGAN.
The generator learns to generate a certain class of the real distribution correlated with a given one-hot format for the latent variables $c$. To obtain this, they proposed to use an inference encoder network capable of performing the mapping $E:\mathcal{X}\rightarrow\mathcal{Z}$, which is the inverse of the generator's mapping and similar to an encoder's mapping for an autoencoder architecture. After the training, the encoder can be employed to classify real data samples according to the latent variable to which it is mostly correlated, thus providing the clustering. {\color{black} Two key differences of our work is that (i) we use a separate generator (not a discrete latent variable) to encode a cluster, enabling our method to discover clusters with more representation capacity, and that (ii) we obtain hierarchical clusters, unlike the previous methods.}

{\color{black}\citet{gan_tree} propose the GAN-Tree framework, which slightly resembles our approach, since it also involves a hierarchical structure of independent nodes containing GANs capable of generating samples related to different levels of a similarity hierarchy. There are several differences, however. The most important is that the main motivation for GAN-Tree was a framework capable of addressing the trade-off between quality and diversity when generating samples from multi-modal data. The authors claimed that their approach could be readily adapted for clustering tasks, but no definitive experiments with clustering benchmarks were provided. Other important difference lies in their splitting procedure, which was performed with a latent $\mathbf{\hat{z}}$ inferred by an encoder $E$ for a sample image $\mathbf{x}$, that is,  $\mathbf{\hat{z}}=E(\mathbf{x})$. For each node of the tree, they decompose their prior for $\mathbf{\hat z}$ into a mixture of two Gaussians with shifted means. They determine the prior Gaussian component to which $\mathbf{\hat{z}}$ is more likely related, and then train the encoder to maximize the likelihood to this prior. For a clustering task, this approach would heavily rely on the assumption that the inference made by $E$, as well as the cluster encoding with the decomposed Gaussians in $\mathcal{Z}$, will be sufficient to capture semantically meaningful clustering patterns. The split in our approach, on the other hand, is directly embedded into the GAN training, with each generator automatically learning to represent each cluster. Therefore, in our work the clustering semantic quality is directly tied to the GAN's well known representation learning capacity, and, in particular, to the tendency of different generators in MGANs to cover different areas of the training distribution with high semantic discrepancy. 
}
\section{Proposed Method: HC-MGAN}

First, we provide a bird's eye view of the proposed method. Then we describe how its two phases -- Raw Split and Refinement -- work in detail.

 \subsection{Overview of the Hierarchical Scheme}

\begin{figure}[t]
\centering
\includegraphics[width=0.99\columnwidth]{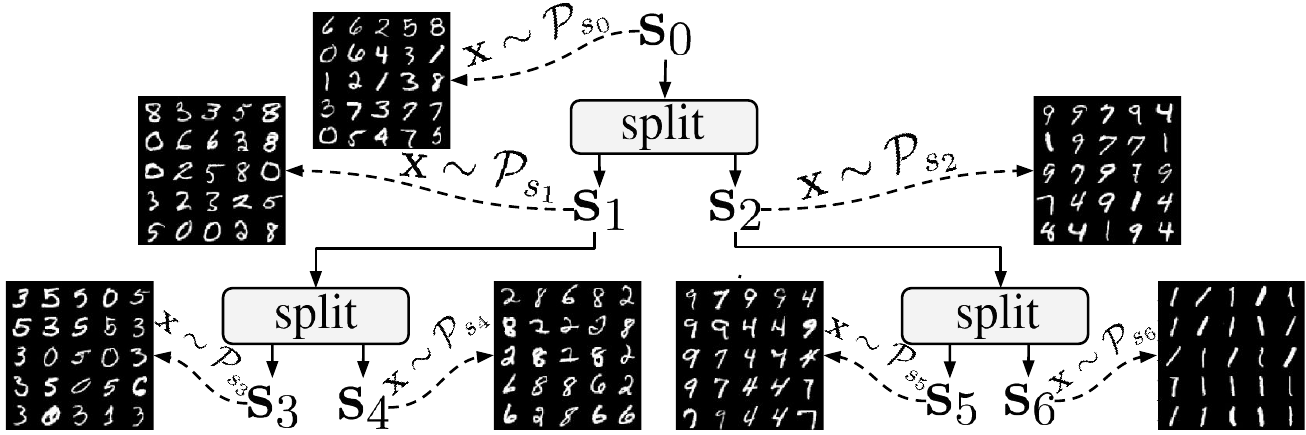}
\caption{{\color{black}Hierarchical clustering tree overview on MNIST. Nodes are represented by membership vectors $\mathbf{s}_k$. Grids show images sampled from each distribution $\mathcal{P}_k$. Each split divides the probability masses in $\mathbf{s}_k$ into two new nodes.}
}
\label{fig:tree_overview_mnist}
\end{figure}

{\color{black}
For a given a collection of $N$ training examples $\mathrm{X_{Data}}=\{\mathbf{x}_1, \mathbf{x}_2, \ldots, \mathbf{x}_N \}$, our method constructs a binary tree of hierarchical clusters, iteratively creating nodes, from top to bottom. The $k$-th created node is represented by a vector $\mathbf{s}_k \in \mathbb{R}^N$, referred to as membership vector, where each $s_{k,i} = p(Z_i=k \mid \mathbf{x}_i, \boldsymbol{\theta})$ measures the probability of example $i$ belonging to cluster $k$ given $\mathbf{x}_i$ and our model's parameters $\boldsymbol{\theta}$, \textit{i.e.}, a soft clustering approach. The initial $\mathbf{s}_0$ consists of an all-ones vector. Figure~\ref{fig:tree_overview_mnist} depicts the initial development of the tree constructed with the MNIST dataset, for the first 7 nodes.} The tree grows via a split mechanism: a soft clustering operation that takes as input a node $\mathbf{s}_k$, 
and divides its probability masses into two new nodes represented by {\color{black} membership vectors $\mathbf{s}_l$ and $\mathbf{s}_m$, with $\mathbf{s}_l + \mathbf{s}_m = \mathbf{s}_k$.} 
We decide which leaf to split next by taking the node $\mathbf{s}_k$ with the largest total mass, since this roughly measures how many examples are associated with it.

{\color{black}
The split mechanism consists of two phases: (i) \textbf{raw split} and (ii) \textbf{refinement phase}, 
as shown in Algorithm~\ref{alg:split}}. Before we obtain the final $\mathbf{s}_l$ and $\mathbf{s}_m$ vectors, vector $\mathbf{s}_k$ undergoes a raw split, which outputs the initial membership vectors $\mathbf{s}^{(0)}_l$ and $\mathbf{s}^{(0)}_m$. 
In most cases, $\mathbf{s}^{(0)}_l$ and $\mathbf{s}^{(0)}_m$ are just rough estimates of how to split $\mathbf{s}_k$ in two clusters. Hence, we use the refinement phase to get them progressively closer to what we expect the ideal soft clustering assignment to be. In Algorithm~\ref{alg:split}, we can see how the refinement transforms $\mathbf{s}^{(0)}_l$ and $\mathbf{s}^{(0)}_m$ into two new membership vectors $\mathbf{s}^{(1)}_l$ and $\mathbf{s}^{(1)}_m$. This process is repeated for $T$ refinement operations to yield the final result $\mathbf{s}_l = \mathbf{s}^{(T)}_l$ and $\mathbf{s}_m = \mathbf{s}^{(T)}_m$. Note that $\mathbf{s}^{(t)}_m + \mathbf{s}^{(t)}_l = \mathbf{s}_k$ for every $t$.

\begin{algorithm}[tb]
\small
\caption{Split}
\label{alg:split}
\textbf{Input}: $\mathrm{X_{Data}}$, $\mathbf{s}_k$
\begin{algorithmic}[1] 

    \STATE $\mathbf{s}_{l}^{(0)},\mathbf{s}_{m}^{(0)}   \leftarrow \mathrm{raw\_split}(\mathrm{X_{Data}}, \mathbf{s}_k) $


   \FOR{$t = 1,\ldots,T$}

    \STATE $\mathbf{s}_{l}^{(t)},\mathbf{s}_{m}^{(t)}   \leftarrow \mathrm{refinement}(\mathrm{X_{Data}}, \mathbf{s}_{l}^{(t-1)},\mathbf{s}_{m}^{(t-1)})$
    \ENDFOR

    
    \RETURN  $\mathbf{s}_{l} = \mathbf{s}_{l}^{(T)},\mathbf{s}_{m} = \mathbf{s}_{m}^{(T)}$

\end{algorithmic}
\end{algorithm}

\begin{figure}[t]
\begin{center}
 \rule{0.9\linewidth}{0pt}
   \includegraphics[width=01.0\linewidth]{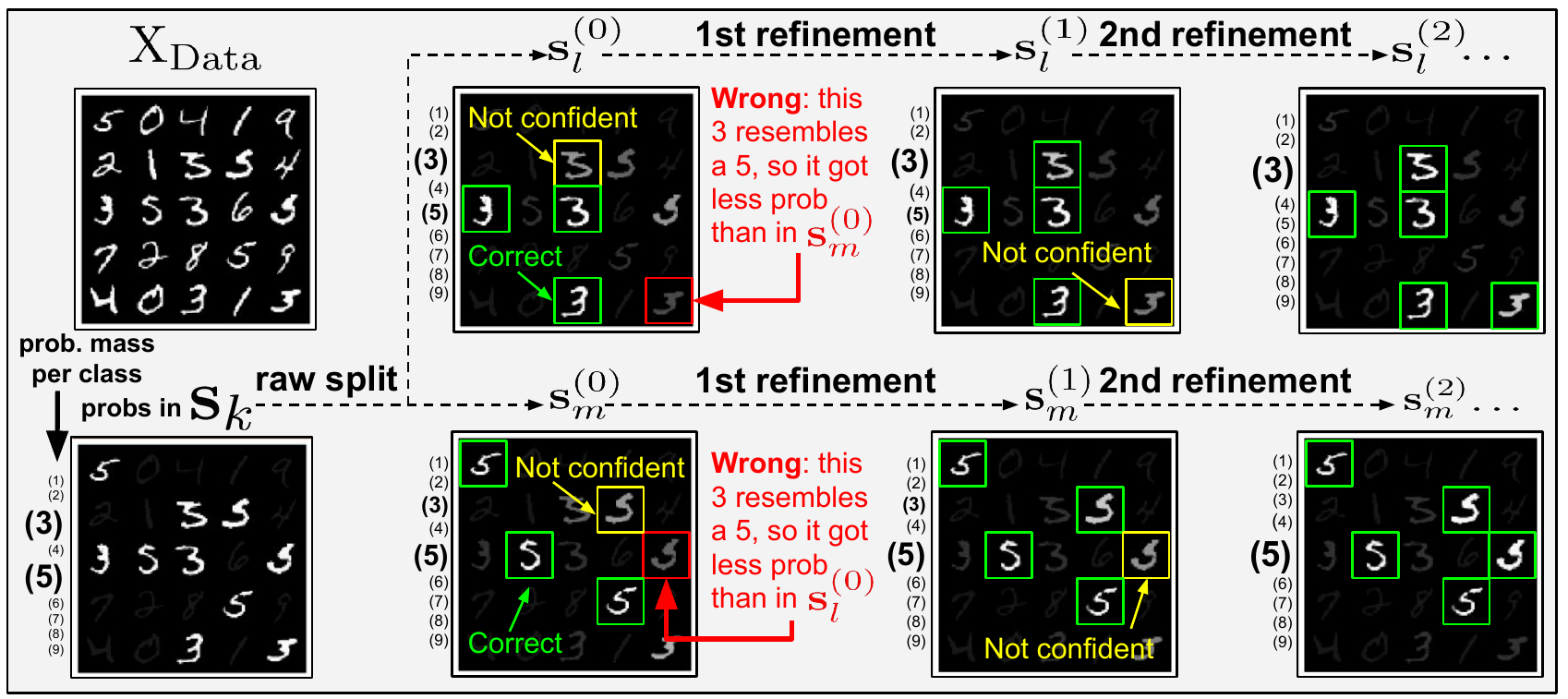}
\end{center}
   \caption{(Best viewed in color) Split mechanism applied to running example $\mathbf{s}_k$ from MNIST. Grids show samples color-coded (gray scale) according to their probability mass in each membership vector (white: 100\%). Keys alongside each grid use a font size scale to indicate mass associated with each class. $\mathbf{s}_k$ is mostly associated with 3's and 5's. Raw split provides a rough separation of two classes. Refinement iterations greatly improve the separation quality.}

\label{fig:split_encapsulated_example}
\end{figure}

Figure~\ref{fig:split_encapsulated_example} shows the progressive improvement resulting from the refinement phase with a grid of 25 MNIST samples, for an $\mathbf{s}_k$ used as a running example. Samples of each membership vector are shown below its label and color-coded (gray scale) according to their probability mass. To visualize how the true class separation changes over iterations, we add a key beside each grid using a font size scale to indicate the amount of probability mass associated with each true class. 

In this example, 3's and 5's are the classes mostly associated with $\mathbf{s}_k$. We expect the final split result ($\mathbf{s}_l$ and $\mathbf{s}_m$) to be a separation of the probability mass of 3's and 5's. After the raw split of $\mathbf{s}_k$, the probability mass of 3's and 5's is roughly divided between $\mathbf{s}_l^{(0)}$ and $\mathbf{s}_m^{(0)}$, but the most ambiguous examples received low probability mass in the membership vector mostly associated to its class. After the first refinement, we observe that $\mathbf{s}_l^{(1)}$ and $\mathbf{s}_m^{(1)}$ provide a better separation of 3's and 5's. After another iteration, most of the probability mass of the 3's (5's) samples is in $\mathbf{s}_l^{(2)}$ ($\mathbf{s}_m^{(2)}$). 


\subsection{Raw Split}
\label{subsection:raw_split}

For the raw split, we use a two-generator MGAN architecture, which we adapt for binary clustering by leveraging the fact that each generator learns to specialize in generating samples from one sub-region of the real data distribution, typically correlated with a specific set of classes of the data.



\begin{figure}[t]
\begin{center}
 \rule{0.9\linewidth}{0pt}
   \includegraphics[width=01.0\linewidth]{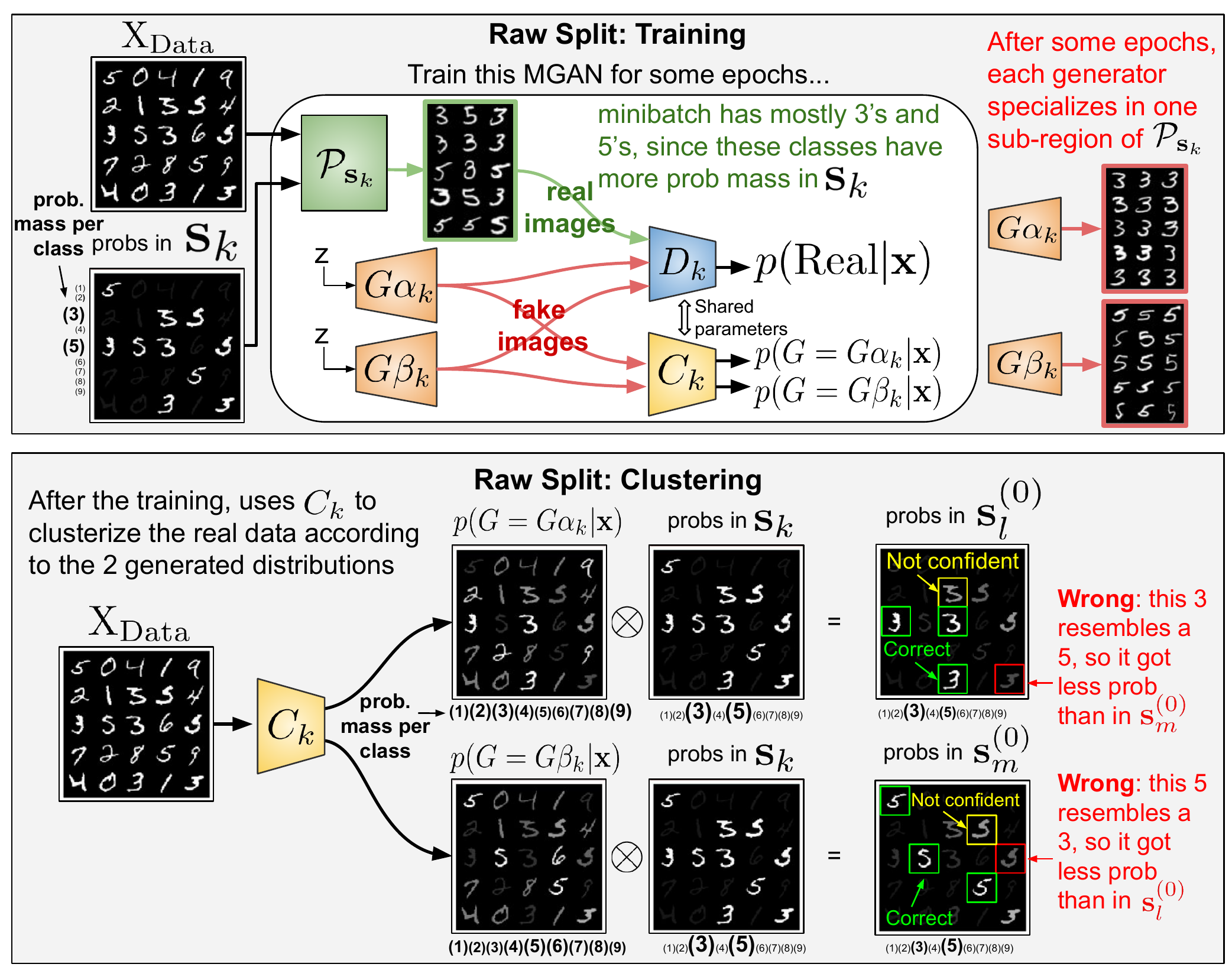}
\end{center}
   \caption{(Top) Training the Raw Split Components: generators $G\alpha_k$, $G\beta_k$, discriminator $D_k$ and classifier $C_k$. $\mathcal{P}_{\mathbf{s}_k}$ draws real samples in proportion to their mass in $\mathbf{s}_k$. (Bottom) Clustering the dataset with the raw split classifier.}

\label{fig:raw_split_compact_vertical}
\end{figure}

Figure \ref{fig:raw_split_compact_vertical} {\color{black}(Top)} depicts the training of this MGAN for our running example. The MGAN components are: generators $G\alpha_k$, $G\beta_k$, discriminator $D_k$ and classifier $C_k$. 
We need each generator to specialize in sub-regions of $\mathbf{s}_k$. Hence, real data samples used for training them must reflect 
the sample distribution given by $\mathbf{s}_k$ (in the example, these are mostly 3's and 5's). We define distribution $\mathcal{P}_{\mathbf{s}_k}$ over the real examples by (sum-to-one) normalizing $\mathbf{s}_k$, and use it to sample the training batches.  
We train the MGAN with the usual adversarial game between generators $G\alpha_k$, $G\beta_k$ and the discriminator $D_k$, while $C_k$ is trained to distinguish between generators. After a few epochs, we observe that one generator is generating mostly 3's while the other, mostly 5's. An important detail is that we also train the generators to minimize $C_k$'s classification loss, increasing the incentive for $G\alpha_k$ and $G\beta_k$ to generate samples from different sub-regions. Moreover, we share some parameters between $D_k$ and $C_k$, since it is more desirable for $C_k$ to perform its classification in a higher-level feature space, such as the one learned by $D_k$.



After training the MGAN for enough epochs, we can cluster the data as shown in Figure \ref{fig:raw_split_compact_vertical} {\color{black}(Bottom)}. We use $C_k$ to perform the clustering on the real data according to the two generated distributions it learned to identify, and which we expect to correlate with different classes of the dataset. In this example, the first generated distribution resembled 3's, while the other resembled 5's. Hence, we expect $C_k$ to mostly assign real 3's probability mass to the first soft cluster and real 5's probability mass to the second. As seen in Figure \ref{fig:raw_split_compact_vertical} {\color{black}(Bottom)}, $C_k$ does manage to roughly assign the 3's (resp.\ 5's) mass to the cluster related to $G\alpha_k$'s (resp.\ $G\beta_k$).  

Note that $C_k$ must be used to perform inference for all examples of every class in the dataset, regardless if they were present in the generated distributions, with $p(G=G\alpha_k|\mathbf{x}) + p(G=G\beta_k|\mathbf{x})=1$. While $C_k$ can be seen as classifying examples conditioned on them belonging to node's $\mathbf{s}_k$ subtree, membership vectors actually correspond to estimates for unconditional probabilities. 
To enforce the condition that, for each example, the sum of its probability masses in the membership vectors of $\mathbf{s}_k$'s children equals vector $\mathbf{s}_k$, we multiply the probabilities of each example predicted by $C_k$ by its probability in $\mathbf{s}_k$.

Finally, by noting that some samples in the running example were not well separated in $\mathbf{s}_l^{(0)}$ and $\mathbf{s}_m^{(0)}$ by classifier $C_k$, we conclude that the two distributions obtained from a raw split may not be sufficiently diverse or have enough quality to account  for the entire set of 3's and 5's the MGAN had access to. This underlines the need for the refinement phase, which will be covered in the next section.

We now provide a formal definition of the MGAN game occurring in the raw split phase. 
Dropping the subscript $k$ to avoid clutter, we define the objective function of the two-generator MGAN 
a sum of two cost functions $\mathcal{L}_{adv}$ and $\mathcal{L}_{cls}$,
\begin{equation}
\footnotesize
\min_{{G\alpha},{G\beta},{C}}\max_{{D}} \mathcal{L} = 
\mathcal{L}_{adv}({G\alpha}, {G\beta}, {D})  +\lambda\mathcal{L}_{cls}({G\alpha},  {G\beta}, C), \label{eq:total_cost_raw_split} 
\end{equation}
where $\mathcal{L}_{adv}$ is the cost for the adversarial minimax game between generators and discriminator, given by 
\begin{equation}
\footnotesize
\label{eq:adv}
\begin{aligned}
&\mathcal{L}_{adv}({G\alpha}, {G\beta}, D) = \,  \mathbb{E}_{\mathbf{x} \sim \mathcal{P}_{\mathbf{s} }}[\mathrm{log} D(\mathbf{x})]\\
&+ \mathbb{E}_{\mathbf{x} \sim \mathcal{P}_{G\alpha}}[\mathrm{log} (1 - D(\mathbf{x}))] + \mathbb{E}_{\mathbf{x} \sim \mathcal{P}_{G\beta}}[\mathrm{log} (1 - D(\mathbf{x}))]
\end{aligned}
\end{equation}
and $\mathcal{L}_{cls}$ is the classification cost that is minimized by both generators and classifier, given by
\begin{equation}
\footnotesize
\label{eq:cls}
\mathcal{L}_{cls}({G\alpha},  {G\beta}, C) =   \mathbb{E}_{\mathbf{x} \sim \mathcal{P}_{G\alpha}}[\mathrm{log} (C(\mathbf{x}))]  + \mathbb{E}_{\mathbf{x} \sim \mathcal{P}_{G\beta}}[\mathrm{log} (C(\mathbf{x}))].
\end{equation}
Note that we multiply $\mathcal{L}_{cls}$ by a regularization parameter $\lambda$ to weight its impact on the total cost. 

Algorithm \ref{alg:raw_split} lists the steps involved in training the MGAN for the raw split and the clustering performed with $C_k$.

\begin{algorithm}[tb]
\small
\caption{Raw Split}
\label{alg:raw_split}
\textbf{Input}: $\mathrm{X_{Data}}$, $\mathbf{s}_k$
\begin{algorithmic}[1] 
\STATE Creates Components $G\alpha_k$, $G\beta_k$, $C_k$, $D_k$\\
\FOR{training\_iterations}

    \STATE sample $\mathbf{x}, \mathbf{\hat{x}}_{\alpha}, \mathbf{\hat{x}}_{\beta}$  from $\mathcal{P}_{\mathbf{s}_k}, \mathcal{P}_{G\alpha_k}, \mathcal{P}_{G\beta_k}$
    
    \STATE Get  $\mathcal{L}_{D_k}^{(real)}$ using $D_k$ criterion on $\mathbf{x}$ with real labels
    
    \STATE Get  $\mathcal{L}_{D_k}^{(fake)}$ using $D_k$ criterion on $\mathbf{\hat{x}}_{\alpha}, \mathbf{\hat{x}}_{\beta}$ w/ fake labels

    \STATE Update $\boldsymbol{\theta}_{D_k}$ with Adam( $\nabla_{\boldsymbol{\theta}_{D_k}} (\mathcal{L}_{D_k}^{(real)} + \mathcal{L}_{D_k}^{(fake)}))$ 
    
    \STATE Get  $\mathcal{L}_{C_k}$ using $C_k$  criterion on $\mathbf{\hat{x}}_{\alpha}, \mathbf{\hat{x}}_{\beta}$, with categorical labels for $G\alpha, G\beta$
    
    \STATE Update $\boldsymbol{\theta}_{C_k}$ with Adam ($\nabla_{\boldsymbol{\theta}_{C_k}} (\mathcal{L}_{C_k})$)

    \STATE Get $\mathcal{L}_{G_k}^{(disc)}$ using $D_k$ criterion on $\mathbf{\hat{x}}_{\alpha}, \mathbf{\hat{x}}_{\beta}$ with real labels
    
    \STATE Get $\mathcal{L}_{G_k}^{(clasf)}$ using $C_k$ criterion on $\mathbf{\hat{x}}_{\alpha}, \mathbf{\hat{x}}_{\beta}$ with categorical labels for $G\alpha, G\beta$
    
    \STATE Update $\boldsymbol{\theta}_{G\alpha_k, G\beta_k}$ w/ Adam( $\nabla_{\boldsymbol{\theta}_{G_k}} (\mathcal{L}_{G_k}^{(disc)} + \lambda\mathcal{L}_{G_k}^{(clasf)} )$)  
\ENDFOR


   \FOR{$\mathbf{x}_i$ in $\mathrm{X_{Data}}$}
    
    \STATE$s_{l,i} \leftarrow C^{(\alpha\_out)}_k(\mathbf{x}_i)\cdot s_{k,i}, \: s_{m,i} \leftarrow C^{(\beta\_out)}_k(\mathbf{x}_i)\cdot s_{k,i}$
    

    \ENDFOR
    \RETURN $\mathbf{s}_{l}$, $\mathbf{s}_{m}$

\end{algorithmic}
\end{algorithm}

\subsection{Refinement}
\label{subsection:refinement_phase}

\begin{figure}[t]
\begin{center}
 \rule{0.9\linewidth}{0pt}
   \includegraphics[width=01.0\linewidth]{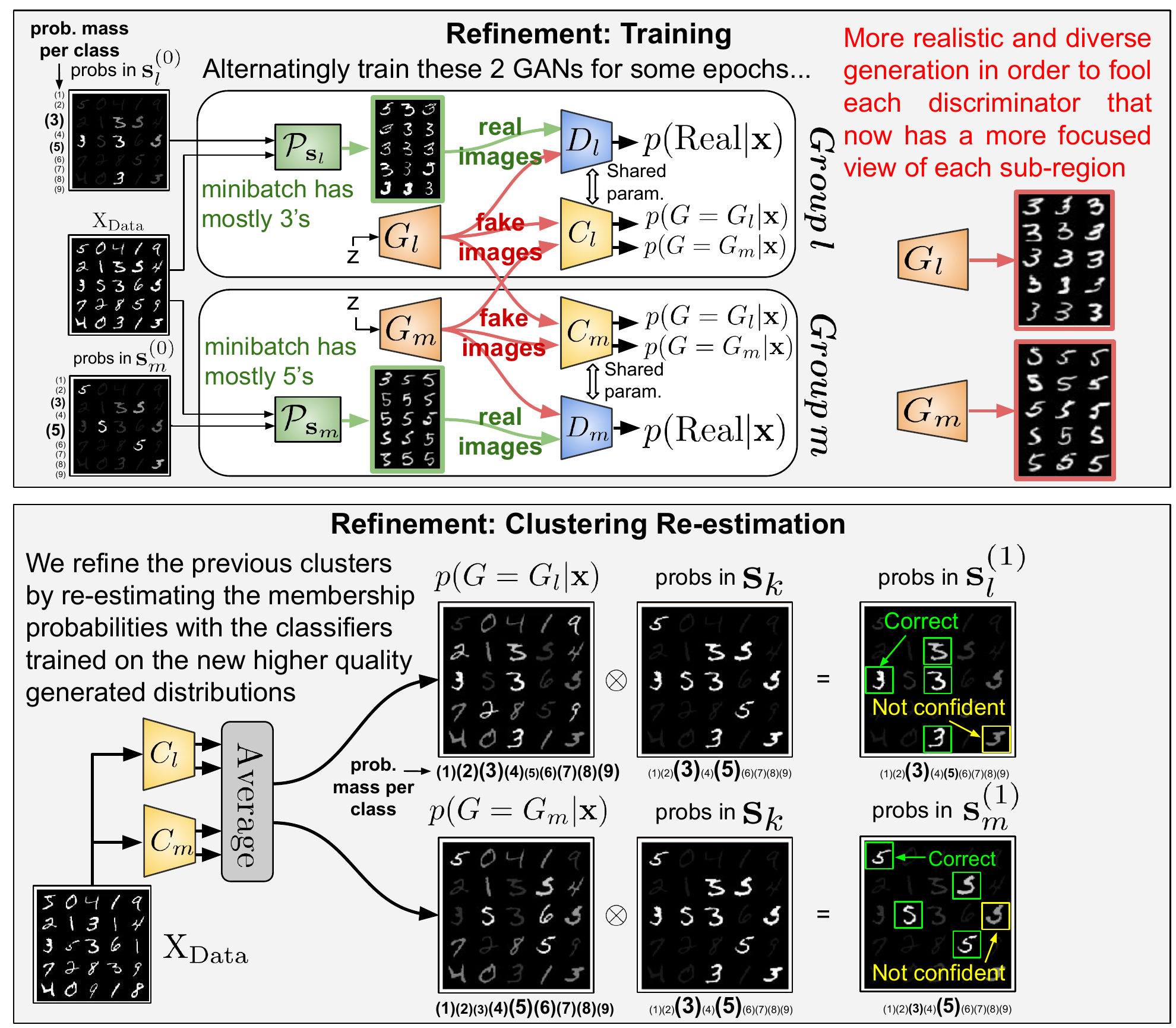}
\end{center}
   \caption{(Top) Refinement Training: generators $G_l$, $G_m$, discriminators $D_l, D_m$ and classifiers $C_l, C_m$. (Bottom) Clustering the dataset with the refinement classifiers. \eat{{\color{red}(Dan) essa legenda tá meio incompleta em comparação à figura com o esquema do raw split. Falta corrigir a um detalhe na na imagem do grupo m.}}}

\label{fig:refinement_compact_vertical}
\end{figure}

The refinement phase comes immediately after the raw split. During this phase, some of the probability mass in the two membership vectors obtained by the raw split, $\mathbf{s}^{(0)}_l$ and $\mathbf{s}^{(0)}_m$, 
is exchanged so as to iteratively improve the clustering quality (see Figure~\ref{fig:split_encapsulated_example}). Each iteration is referred to a refinement sub-block.
%
%
Without loss of generality, consider the first refinement sub-block, whose components are depicted in Figure \ref{fig:refinement_compact_vertical} {\color{black}(Top)}, for our running example.
The components are divided in two ``refinement groups'', $l$ and $m$ (each formed by a generator $G_i$, a discriminator $D_i$ and a classifier $C_i$, $i \in \{l,m\}$). Each group $i$ takes $\mathbf{s}^{(0)}_i$ as input, and has its own independent GAN game occurring between  $G_i$ and $D_i$. 

This scheme with two separated GANs is designed to obtain a more focused generative representation of each sub-region of $\mathbf{s}_k$ 
than we were able to obtain at the raw split phase using a single MGAN's discriminator to learn to discriminate the entire region described {\color{black}by} $\mathbf{s}_k$. \textit{By providing a more focused view of one sub-region to one 
discriminator, it encounters less variance among the real examples it receives, and thus its discriminative task becomes easier. We expect its adversarial generator's response to be a more diverse and convincing generation of examples associated with that particular sub-region.
}

As shown in Figure~\ref{fig:refinement_compact_vertical} {\color{black}(Top)}, each GAN in groups $i \in \{l, m\}$ draws real samples from its corresponding distribution $\mathcal{P}_{\mathbf{s}_i}$ over $\mathrm{X_{Data}}$, which is equal to the (sum-to-one) normalized vector $\mathbf{s}_i$ (analogously to $\mathcal{P}_{\mathbf{s}_k}$). As a result, the minibatches passed to each group's GAN reflect the probability mass in their respective membership vectors, \textit{e.g.}: in our running example, $\mathcal{P}_{\mathbf{s}_l}$ draws mostly 3's, since 3's have more probability mass in $\mathbf{s}^{(0)}_l$, but it might eventually draw some 5's as well, since there's still some mass for 5's in $\mathbf{s}^{(0)}_l$.



The role of classifiers $C_l$ and $C_m$ in these two separated GAN games is similar to the single classifier $C_k$ used in the raw split phase: learn to distinguish samples from $G_l$ and $G_m$, thereby providing a way to cluster the real data. However, instead of using a single classifier (as in the raw split), we found that using two separate classifiers which share parameters with the respective discriminators forces the classification to occur in a higher-level feature space, achieving better clustering. Furthermore, it allows us to train $G_l$ and $G_m$ to minimize the classifiers' losses, thus providing incentive for each generator to generate samples more strongly correlated with each sub-region (we did something similar for the 2 generators in the MGAN of the raw split).



After training the two refinement groups alternately for enough epochs, we perform a \textbf{clustering re-estimation}, as shown in Figure~\ref{fig:refinement_compact_vertical} {\color{black}{(Bottom)}}. This is similar to the way $C_k$ was used to cluster the real data in the raw split (Figure~\ref{fig:raw_split_compact_vertical} {\color{black}(Bottom)}). Now, both classifiers estimate the probability that a sample $\mathbf{x}_i \in \mathrm{X}_\mathrm{Data}$ came from $G_l$ (or its complement, $G_m$), hereby denoted by $C_i^{(l\_out)}(\mathbf{x}_i)$, for $i \in \{l,m\}$.  
We take the average probability \eat{{\color{red}Dan: acho que esse valor médio deveria ir no texto, sem centralizar. Ta ocupando 3 linhas.}}
$(C_l^{(l\_out)}(\mathbf{x}_i) + C_m^{(l\_out)}(\mathbf{x}_i))/2$
as the proportion of the mass associated with $\mathbf{x}_i$ in $\mathbf{s}_k$ that should go into $\mathbf{s}_l^{(1)}$ and. By using this ``two-classifier ensemble'', we aim to increase the quality of the clustering, since the generators' distributions are expected to be more representative of each sub-region of the dataset than the two generated distributions obtained during the raw split.  
In the running example, $G_l$'s ($G_m$'s) distributions resembled 3's (5's), so the classifiers assign more of the 3's (5's) probability mass to the cluster associated with $G_l$ ($G_m$'s). 

For the subsequent refinement, we expect that using $\mathbf{s}_l^{(1)}$ and $\mathbf{s}_m^{(1)}$ to train new refinement groups $l$ and $m$ can yield generated distributions that are even more representative of the sub-regions encoded by these membership vectors, providing, in turn, even more information for the classifiers to perform the clustering and to obtain improved membership vectors $\mathbf{s}_l^{(2)}$ and $\mathbf{s}_m^{(2)}$ (recall Figure \ref{fig:split_encapsulated_example}). Therefore, by repeating the process over $T$ refinements, it is expected that the initial sub-regions captured by $\mathbf{s}_k$ are increasingly more associated with either of the refinement groups.

We now provide a formal definition for the two simultaneous GAN games occurring for the training of the components of the refinement phase. From the perspective of refinement group $l$, the training can be defined as an optimization of a sum of two cost functions $\mathcal{L}_{adv}$ and $\mathcal{L}_{cls}$,
\begin{equation}
\footnotesize
\label{eq:total_cost_refinement}
\min_{{G_l}
,{C_l}}\max_{{D_l}} \mathcal{L}(G_l, D_l, C_l) = \mathcal{L}_{adv}(G_l, D_l) +\lambda\mathcal{L}_{cls}(G_l, C_l),
\end{equation}
where $\mathcal{L}_{adv}$ describes the cost function for the adversarial minimax game between generator $G_{l}$ and discriminator $D_{l}$, that only involves group $l$ components, and is given by
\begin{equation}
\footnotesize
\label{eq:adv_ref}
\mathcal{L}_{adv}(G_{l},D_{l})  = \,  \mathbb{E}_{\mathbf{x} \sim \mathcal{P}_{data }}[\mathrm{log} D_{l}(\mathbf{x})]
 + \mathbb{E}_{\mathbf{x} \sim \mathcal{P}_{G_{l}}}[\mathrm{log} (1 - D_{l}(\mathbf{x}))]
\end{equation}
and $\mathcal{L}_{cls}$ is classification cost that is minimized with respect to $G_{l}$'s parameters and $C_{l}$'s parameters, but also involves $G_{m}$ and $C_{m}$ for computing the cost, given by
\begin{equation}
\footnotesize
\label{eq:cls_ref}
\begin{aligned}
\mathcal{L}_{cls}(G_{l},C_{l}) &= \, \mathbb{E}_{\mathbf{x} \sim \mathcal{P}_{G_{l}}}[\mathrm{log} \, C_{l}(\mathbf{x})] + \mathbb{E}_{\mathbf{x} \sim \mathcal{P}_{G_{l}}}[\mathrm{log} \, C_{m}(\mathbf{x})] \\
& + \mathbb{E}_{\mathbf{x} \sim \mathcal{P}_{G_{m}}}[\mathrm{log} \, C_{l}(\mathbf{x})].
\end{aligned}
\end{equation}
We multiply $\mathcal{L}_{cls}$ by a regularization parameter $\lambda$ to weight its impact on the total cost. The corresponding equations for refinement group $m$ follow analogously. 

Algorithms \ref{alg:refinement} and  \ref{alg:train_refinement_group} respectively list the steps involved in the external training loop that coordinates the alternated training of refinement groups $l$ and $m$, and in the function called by Algorithm~\ref{alg:refinement} that performs an update for the components of a given refinement group isolatedly.


\begin{algorithm}[tb]
\small
\caption{Refinement}
\label{alg:refinement}
\textbf{Input}: $\mathrm{X_{Data}}, \mathbf{s}^{(t)}_l \mathbf{s}^{(t)}_m$

\begin{algorithmic}[1] 
\STATE Creates $G_l, D_l, C_l$ for group $l$ and $G_m, D_m, C_m$ for group $m$\\

   \FOR{$T$ iterations}
    
    \STATE sample $\mathbf{x}_l, \mathbf{x}_m, \mathbf{\hat{x}}_l, \mathbf{\hat{x}}_m$  from $\mathcal{P}_{\mathbf{s}^{(t)}_l}, \mathcal{P}_{\mathbf{s}^{(t)}_m}, \mathcal{P}_{G_l}, \mathcal{P}_{G_m}$

    \STATE TrainRefinGroup($\mathrm{\mathcal{G}}_{int}=\{G_l, D_l, C_l, \mathbf{x}_l, \mathbf{\hat{x}}_l\}, $\\
    $ \mathrm{\mathcal{G}}_{ext}=\{C_m,\mathbf{\hat{x}}_m\}$) \COMMENT{trains \textit{l} with needed external data/components from \textit{m}}

    \STATE TrainRefinGroup($\mathrm{\mathcal{G}}_{int}=\{G_m, D_m, C_m, \mathbf{x}_m, \mathbf{\hat{x}}_m\},$\\
    $ \mathrm{\mathcal{G}}_{ext}=\{C_l, \mathbf{\hat{x}}_l\}$) \COMMENT{trains \textit{m} with needed external data/components from \textit{l}}
    
    \ENDFOR


   \FOR{$\mathbf{x}_i$ in $\mathrm{X_{Data}}$}

    \STATE $s_{i,l}^{(t+1)} \leftarrow  (C^{ (l\_out)}_l(\mathbf{x}_i) + C^{( l\_out)}_m(\mathbf{x}_i))\cdot (s_{i,l}^{(t)}+s_{i,m}^{(t)})/2$
    
    \STATE $s_{i,m}^{(t+1)} \leftarrow  (C^{ (m\_out)}_l(\mathbf{x}_i) + C^{( m\_out)}_m(\mathbf{x}_i))\cdot (s_{i,l}^{(t)}+s_{i,m}^{(t)})/2$

    \ENDFOR
    \RETURN $\mathbf{s}_{l}^{(t+1)}$, $\mathbf{s}_{m}^{(t+1)}$
             
\end{algorithmic}
\end{algorithm}

\begin{algorithm}[tb]
\small
\caption{TrainRefinGroup}
\label{alg:train_refinement_group}
\textbf{Input}: $\mathrm{\mathcal{G}}_{int} = \{G_{int}, D_{int}, C_{int}, \mathbf{x}, \mathbf{\hat{x}}_{int}\}, \mathrm{\mathcal{G}}_{ext}=\{ C_{ext}, \mathbf{\hat{x}}_{ext} \}$

\#$\mathrm{\mathcal{G}}_{int}$ receives internal data/components from the current refinement group being trained, $\mathrm{\mathcal{G}}_{ext}$  receives external data/components from the neighbor refin. group needed to train the current group

\begin{algorithmic}[1] 

    \STATE Get  $\mathcal{L}_{D_{int}}^{(real)}$ using $D_{int}$ criterion on $\mathbf{x}$ with real labels
    
    \STATE Get  $\mathcal{L}_{D_{int}}^{(fake)}$ using $D_{int}$ criterion on $\mathbf{\hat{x}}_{int}$ with fake labels
    
    \STATE Updates $\boldsymbol{\theta}_{D_{int}}$ with  Adam($\nabla_{\boldsymbol{\theta}_{D_{int}}} (\mathcal{L}_{D_{int}}^{(real)} + \mathcal{L}_{D_{int}}^{(fake)})$)

    \STATE Get  $\mathcal{L}_{C_{int}}$ using $C_{int}$  criterion on $\mathbf{\hat{x}}_{int}, \mathbf{\hat{x}}_{ext}$, with categorical labels for internal and external generated data

    \STATE Updates $\boldsymbol{\theta}_{C_{int}}$ with  Adam($\nabla_{\boldsymbol{\theta}_{C_{int}}} (\mathcal{L}_{C_{int}})$)

    \STATE Get $\mathcal{L}_{G_{int}}^{(disc)}$ using $D_{int}$ criterion on $\mathbf{\hat{x}}_{int}$ with real labels
    
    \STATE Get $\mathcal{L}_{G_{int}}^{(clasf)}$ using $C_{int}$ criterion on $\mathbf{\hat{x}}_{int}, \mathbf{\hat{x}}_{ext}$ with categorical labels for internal and external generated data
   
    \STATE Updates $\boldsymbol{\theta}_{G_{int}}$ with  Adam($\nabla_{\boldsymbol{\theta}_{G_{int}}} (\mathcal{L}_{G_{int}}^{(disc)} + \lambda\mathcal{L}_{G_{int}}^{(clasf)} )$)  
             
\end{algorithmic}
\end{algorithm}

\section{Experiments}

We used the same standard convolutional GAN/MGAN architecture for all the following experiments. Respecting the unsupervised nature of clustering, which does not afford hyperparameter tuning, we only selected slightly different tunings for each dataset, none of which required labeled supervision, merely aiming to stabilize the generators and to avoid overfitting with classifiers while distinguishing between generators. The code for HC-MGAN will be available at https://github.com/dmdmello/HC-MGAN.

\subsubsection{Datasets}

{\color{black}
We consider three datasets: MNIST \cite{lecun1998gradient}, Fashion MNIST (FMNIST) \cite{xiao2017fashion} and Stanford Online Products (SOP) \cite{Song_2016_CVPR}. Following a common practice in DC works, we used all available images for each dataset. \textbf{MNIST}: This dataset consists of grayscale images of hand-written digits with 28x28 resolution, with 10 classes and approximately 7k images available for each class. \textbf{FMNIST}: This dataset consists of gray scale pictures of clothing-related objects with 28x28 resolution, with 10 classes and exactly 7k images available for each class. \textbf{SOP}: This dataset consists of color pictures of products with varying resolution sizes, with 12 classes, and a varied number of examples per class, roughly ranging from 6k examples to 13k. SOP, in particular, is designed for supervised tasks, and is very hard for clustering due to high intra-class variance. We follow the practice adopted for SOP in \cite{benchmark:NCSC}, i.e., we convert the images to grayscale, resize them to 32x32 resolution, and drop the classes ``kettle'' and ``lamp'' from it.
}

\subsubsection{Evaluation Metrics}

We consider two of the most common clustering metrics for evaluating our method's clustering performance on each dataset: \textit{clustering accuracy} (ACC) and \textit{normalized mutual information} (NMI). As usual, these metrics are computed on the results obtained when setting the number of clusters $C$ to the number of classes in the data. {\color{black} For a direct comparison, we let HC-MGAN's tree grow until it reaches $C$ leaves. Additionally, we convert the final soft clustering outputs to hard assignments by attributing each example to the highest probability group, and then compute the evaluation metrics.
}

\subsubsection{Baselines and SOTA methods}
\label{subsection:benchmarks}

We present the baselines and state-of-the-art methods used in the comparison.  We consider five groups of methods: (i) classical, non Deep Learning (DL)-based based: K-Means \cite{kmeans}, SC  \cite{benchmark:SC}, AC \cite{benchmark:AC}, NMF \cite{benchmark:NMF}; (ii) Varied DC: DEC \cite{dec}, DCN \cite{dcn}, JULE \cite{benchmark:JULE}, VaDE \cite{benchmark:VaDE}, DEPICT \cite{benchmark:DEPICT}, SpectralNET, DAC \cite{benchmark:DAC},  \cite{benchmark:SpectralNet}, DualAE \cite{benchmark:DualAE}; (iii) Subspace Clustering (either DL-based or not): NCSC \cite{benchmark:NCSC}, SSC \cite{benchmark:SSC}, LLR \cite{benchmark:LLR}, KSSC \cite{benchmark:KSSC}, DSC-Net \cite{benchmark:DSC-Net}, $k$-SCN \cite{benchmark:k-SCN}; (iv) GAN-based DC: ClusterGAN \cite{clustergan}, InfoGAN \cite{infogan}, DLS-Clustering \cite{benchmark:DLS-Clustering};(v) DC w/ data augumentation: \cite{benchmark:NCSC}, IIC \cite{benchmark:IIC}, DCCM \cite{benchmark:DCCM} and DCCS \cite{benchmark:DCCS}. \emph{None of the DC methods we found in the literature are hierarchical.}
{\color{black}Most results are transcribed from either \cite{benchmark:DCCS}, \cite{benchmark:NCSC} or \cite{clustergan}}.

\subsubsection{Results}
Table \ref{tab:main_results} shows the performance comparison between traditional baselines, state-of-the-art DC methods without and with data augumentation, and our method. In order to check the effectiveness of the refinements, we also display results obtained only with raw splits. 

\noindent \textbf{MNIST} Our method outperforms the traditional baselines by a large margin. In terms of ACC, it is not among the top 5 presented methods, but it performs reasonably close to them, even outperforming, in either NMI or ACC, some DC methods like ClusterGAN, InfoGAN, DEC, DCN. This result was obtained by employing the same architecture we used for FMNIST, which might be of excessive capacity for MNIST and even harm the clustering result by making it trivial for a single generator to represent all the data, instead of a cluster of similar data points, during the raw split. Of all the datasets, this was the one for which the refinements showed the greatest improvement over raw split experiment. 

\noindent \textbf{FMNIST} Only DCCS is able to surpass our method's ACC and NMI performance. {\color{black}We emphasize that using data augmentation in DCCS causes a significant improvement, since selecting the right type of augmentations for a specific dataset can reduce much of the intra-class variance. However, augmentation with GANs is challenging \cite{gan_augmentation_challenge}, so we leave this for future work.} The refinements still had a positive impact over the raw split only experiment.

{\color{black}
{\noindent \textbf{SOP} The results for existing methods were transcribed from the NCSC work~\cite{benchmark:NCSC}. There the authors state that they handpicked 1k examples per class to create a manageable dataset for clustering, but did not specify how or which examples were selected. Their choice was made in the context of competing subspace clustering methods to which their model was being compared, many of which are not able to scale to larger datasets due to the memory constraints involved in computing a similarity matrix necessary for their methods.  
We tried without success to contact the authors to obtain the same subset of the data. Therefore, we decided to evaluate our results on the entire data. Due to the class imbalance, we compute both the mean ACC over classes (reported in the table) and the overall ACC achieved by HC-MGAN, which are, respectively, 0.229 and 0.221. NMI is invariant to class imbalance. 
On this dataset, our model does not perform as well as the subspace clustering group, especially NCSC, KSSC and DSC-Net, even though it was on par with NCSC and greatly surpassed the KSSC and DSC-Net on other datasets. But for other DC methods reported from~\cite{benchmark:NCSC}, our method performs closely, even outperforming DCN and InfoGAN in accuracy.}
}

\begin{table}[t]
\begin{center}
\small
\setlength{\tabcolsep}{4pt}
\begin{tabular}{llllllllll}
\toprule
Dataset  & \multicolumn{2}{c}{MNIST} & \multicolumn{2}{c}{FMNIST} & \multicolumn{2}{c}{SOP} \\ \hline
Metrics                  & ACC           & NMI      & ACC       & NMI     & ACC       & NMI     \\ \hline
K-$\mathrm{means}^\mathrm{i}$                       &.572         &.500     &.474       &.512     & -      & -   \\ 
$\mathrm{SC}^\mathrm{i}$                            &.696         &.663     &.508       &.575     & -      & -   \\  
$\mathrm{AC}^\mathrm{i}$                            &.695         &.609     &.500       &.564     & -      & -   \\  
$\mathrm{NMF}^\mathrm{i} $                          &.545         &.608     &.434       &.425     & -      & -   \\  \hline
$\mathrm{DEC}^\mathrm{ii}  $                        &.843         &.772     &.590       &.601     &\textbf{.229}    &.121  \\
$\mathrm{DCN}^\mathrm{ii}  $                        &.833         &.809     &.587       &.594     &.213    &.084 \\  
$\mathrm{JULE}^\mathrm{ii}   $                      &.964         &.913     &.563       &.608     & -      & -      \\  
$\mathrm{VaDE}^\mathrm{ii}  $                       &.945         &.876     &.578       &.630     & -      & -      \\ 
$\mathrm{SpectralNet}^\mathrm{ii}$        &\textbf{.971 } &.924    &.533 &.552     & -      & -      \\    
$\mathrm{DualAE}^\mathrm{ii}$           &\textbf{.978 } &\textbf{.941}    &\textbf{.662}     &.645     & -      & -      \\   
$\mathrm{DAC}^\mathrm{ii}$                          &.966  &\textbf{.967}   &.615     &.632     &\textbf{.231}    &.098  \\  \hline
$\mathrm{SSC}^\mathrm{iii}$                         &.430         &.568    &.359     &.181     &.127    &.007      \\  
$\mathrm{LLR}^\mathrm{iii}$                         &.552         &.665    &.345     &.254     &.224    &\textbf{.173}      \\  
DSC-$\mathrm{Net}^\mathrm{iii}$                     &.659         &.730    &.606     &.617     &\textbf{.269 }   &\textbf{.146 }     \\
$\mathrm{KSSC}^\mathrm{iii}$                        &.585         &.677    &.382     &.197     &\textbf{.268}    &\textbf{.152}      \\
k-$\mathrm{SCN}^\mathrm{iii}$                       &.871         &.782    &.638     &.620     &\textbf{.229}    &\textbf{.166 }     \\ 
$\mathrm{NCSC}^\mathrm{iii}$                        &.941   &.861 &\textbf{.721}     &\textbf{.686 }    &\textbf{.275 }   &\textbf{.138 }     \\   \hline
$\mathrm{ClustGAN}^\mathrm{iv}$                     &.950         &.890    &.630     &.640     & -      & -      \\
$\mathrm{InfoGAN}^\mathrm{iv}$                      &.890         &.860    &.610     &.590     &.198    &.082      \\  
DLS-$\mathrm{Clst}^\mathrm{iv}$       &\textbf{.975}&\textbf{.936 }  &\textbf{.693 }  &\textbf{.669 }    & -      & -      \\    \hline
$\mathrm{IIC}^\mathrm{v}$               &\textbf{.992}&\textbf{.978}  &.657   &.637     & -      & -   \\   
$\mathrm{DCCM}^\mathrm{v}$                          & -          & - &      .657 &.637     & -      & -       \\   
$\mathrm{DCCS}^\mathrm{v}$               &\textbf{.989}  &\textbf{.970}     &\textbf{.756 }    &\textbf{.704}     & -      & -     \\ \hline
Ours                                                &.943        &.905     &\textbf{.721}     &\textbf{.691 }    &\textbf{.229}    &.072           \\ 
Ours (raw)                                          &.877       &.856     &.704     &\textbf{.690  }   &.204    & .063           \\ \hline

\hline
\end{tabular}
\end{center}
\caption{ 
Clustering performance results on 3 datasets w.r.t.\ ACC and NMI (top 5 in bold). ``Ours'' indicates HC-MGAN, with (raw) indicating no refinement operations performed. $^\mathrm{i}$: non-deep. $^\mathrm{ii}$: varied DC methods. $^\mathrm{iii}$: Subspace Clustering. $^\mathrm{iv}$: GAN-based DC. $^\mathrm{v}$: DC w/ data augmentation.}
\label{tab:main_results}
\end{table}

\begin{figure}[t]
    \centering
    \includegraphics[width=0.99\columnwidth]{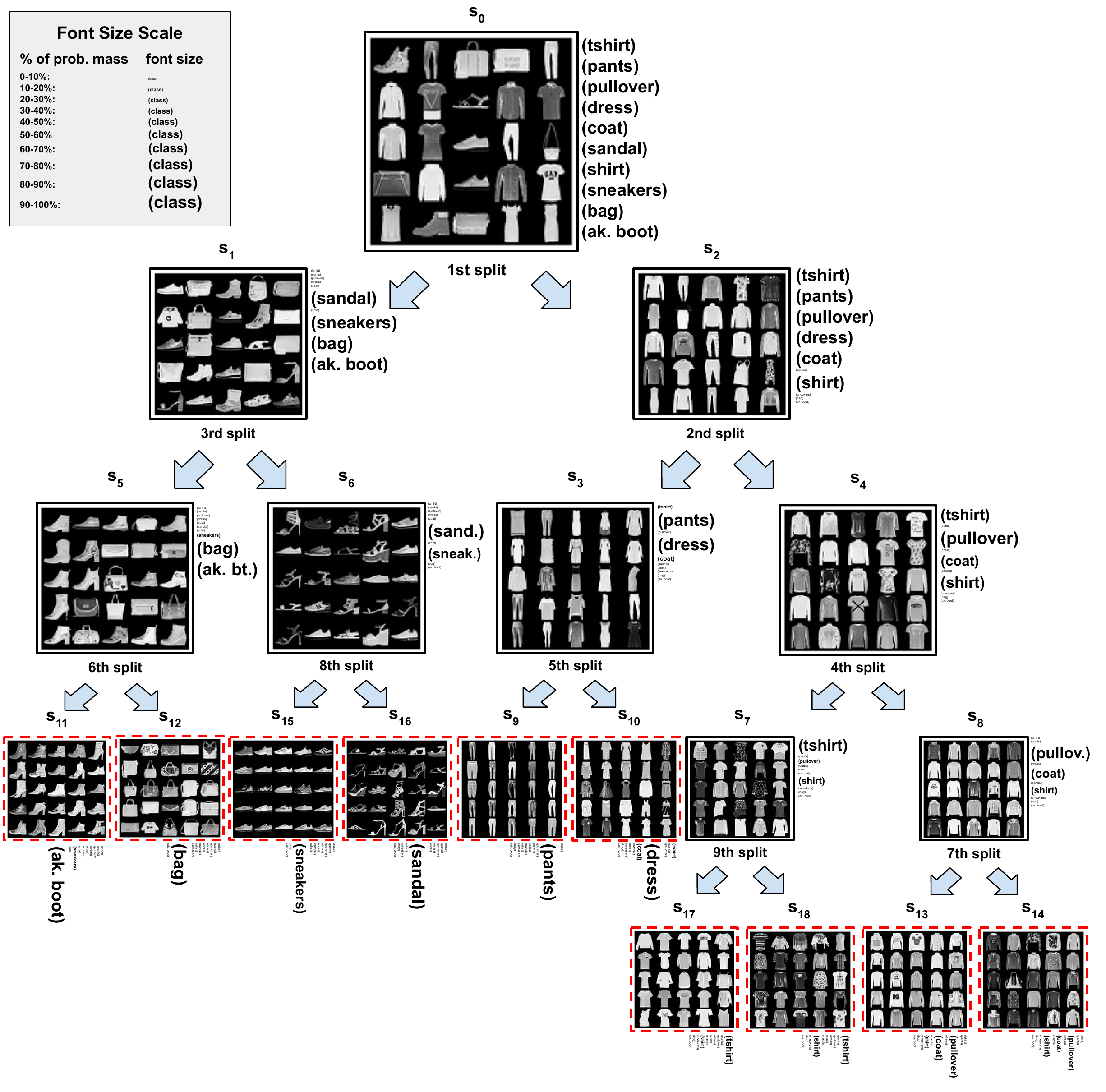}
    \caption{Tree for the FMNIST dataset. Grids show sampled images; keys beside grid represent probability mass of each class in font size scale. Dashed red line indicate leaf nodes. 
    }
    \label{fig:tree_viz_fmnist_size_code}
\end{figure}

\subsubsection{Qualitative Analysis}

Figure~\ref{fig:tree_viz_fmnist_size_code} shows a visualization of the clustering tree for FMNIST. For each node $k$ in the tree, it displays a grid of 25 real examples sampled in proportion to the weights in membership vector $\mathbf{s}_k$. The prevalence of each class in $\mathbf{s}_k$ 
is represented in a font size scale by the class names inside parenthesis beside each grid. Specifically, the probability mass of a class $A$ in $\mathbf{s}_k$ is given by $\sum_{i}^{N} s_{k,i} \cdot \boldsymbol{1}(c_i=A)$, where $\boldsymbol{1}$ is an indicator function and $c_i$ is the class of the $i$-th example. Through a quick inspection, we observe that the \nth{1} split ($\mathbf{s}_0$) occurred with almost perfect precision, with examples of the same class having their probability mass nearly entirely allocated to either $\mathbf{s}_1$ or $\mathbf{s}_2$. We begin to notice some imprecision at the \nth{2} split ($\mathbf{s}_2$), with a small portion of the probability mass of classes coat and t-shirt being assigned to $\mathbf{s}_3$ while the largest portion went to $\mathbf{s}_4$, and at the \nth{3} split  ($\mathbf{s}_1$), with a small portion of the sneakers' mass being assigned to $\mathbf{s}_5$ and the largest portion going to $\mathbf{s}_6$. The most imprecise splits occurred during the \nth{4}, \nth{7} and \nth{9} splits (i.e., $\mathbf{s}_4$, $\mathbf{s}_6$ and $\mathbf{s}_7$) but the classes involved in these splits (t-shirt, pullover, coat and shirt) are the most visually similar in the dataset, thus being the hardest to accurately separate into clusters. The other classes are relatively well separated. One fact that can explain why our method performed very well w.r.t.\ NMI on FMNIST is that most of the probability mass of each class for which our method exhibited low accuracy was assigned to at most two clusters. %
For the NMI metric, having the wrongfully assigned classes concentrated on fewer clusters provides better mutual information than having them spread throughout more clusters.

\section{Conclusion}

In this work, we proposed a method for hierarchical clustering that leverages the deep representation capability of GANs and MGANs. {\color{black}To the best of our knowledge, this is the first hierarchical DC method}. It constructs a tree of clusters from top to bottom, where each leaf node represents a cluster. Each cluster is divided in binary splits, which are performed in two phases: a raw split and a refinement phase. We have shown how well our method compares to other deep clustering techniques on clustering datasets, obtaining competitive results {\color{black}against other DC methods that lack advantages of the hierarchical approach.}  

\section{ Acknowledgments}

The authors thank the partial support from Brazilian research supporting agencies: CNPq (grant PQ 313582/2018-1), FAPEMIG (grant CEX - PPM-00598-17, grant APQ-02337-21), and CAPES (grant 88887.506739/2020-00).  

\bibliography{aaai22}
\section{Additional Results}

\begin{figure*}
\begin{center}
 \rule{0.7\textwidth}{0pt}
   \includegraphics[width=0.9\linewidth]{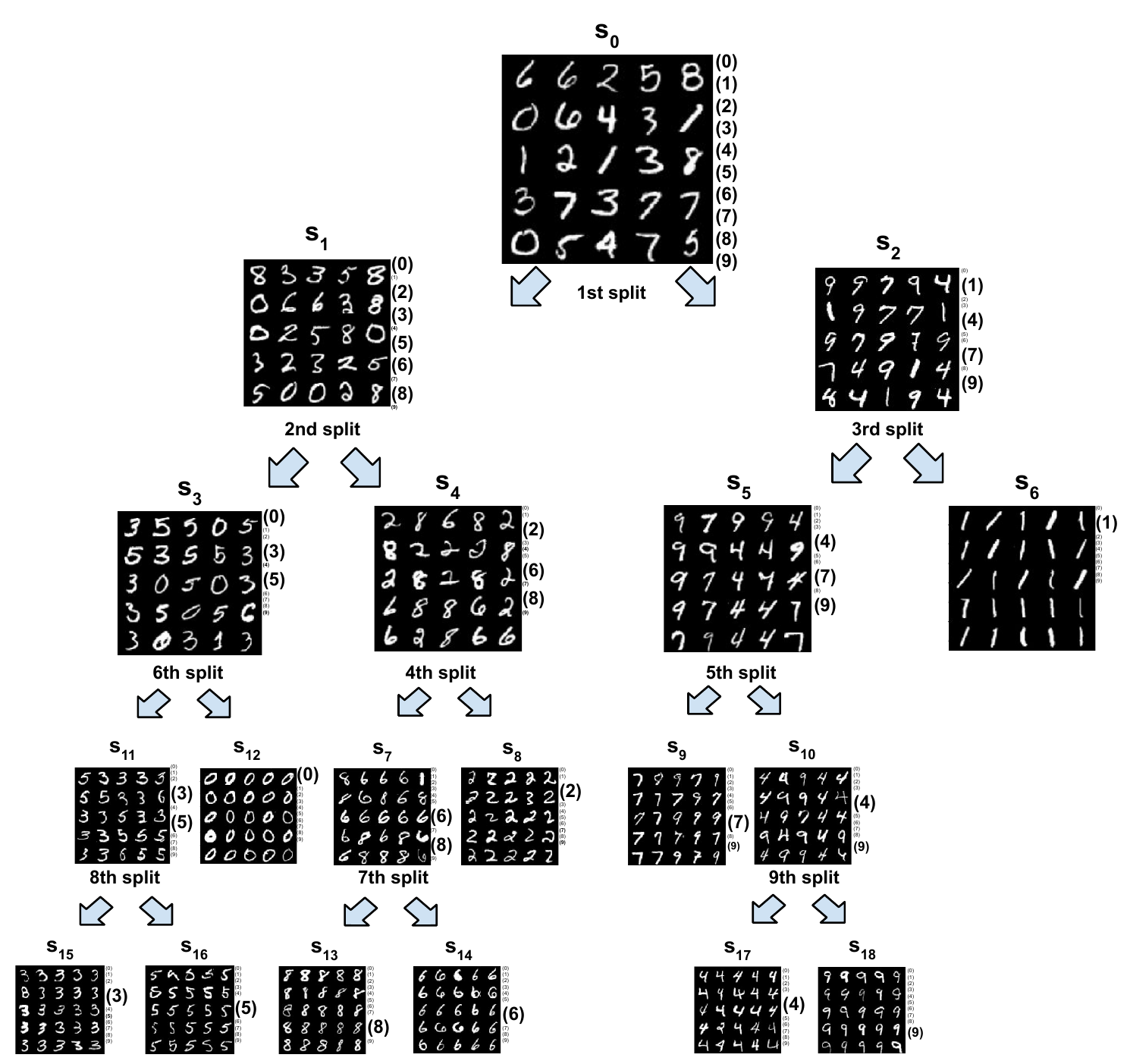}
\end{center}
   \caption{Hierarchical clustering tree for the MNIST dataset. Grids show sampled images; keys beside grid represent probability mass of each class in font size scale. 
}

\label{fig:tree_viz_mnist}
\end{figure*}

\begin{figure*}
\begin{center}
 \rule{0.7\textwidth}{0pt}
   \includegraphics[width=0.9\linewidth]{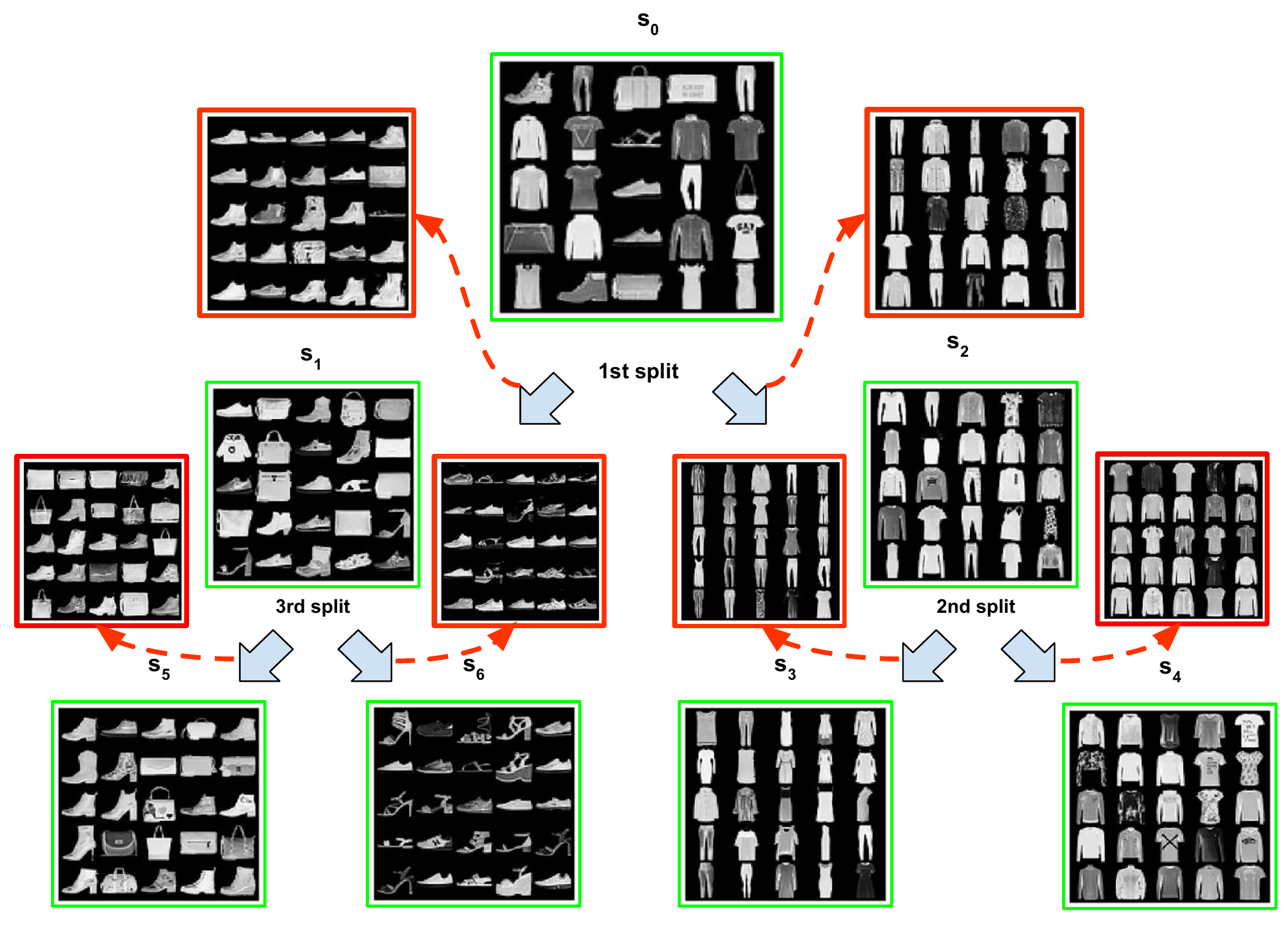}
\end{center}
   \caption{Visualizing the image generation for the first 3 splits on the FMNIST dataset. Red borders indicate samples generated by the generators trained at the last refinement iteration for each node. Green borders indicate real image samples according to each node's probability distribution, reproducing the same samples used for Figure 5 in the main paper. We can clearly observe how the generated samples visually correlate with each clustering pattern obtained, as expected.
}

\label{fig:generators_output_fmnist}
\end{figure*}

In Figure \ref{fig:tree_viz_mnist} we depict an overview of the hierarchical clustering tree obtained with the MNIST dataset, similarly as we did for FMINIST in Figure 5 in the main paper. In Figure \ref{fig:generators_output_fmnist} we depict the images generated at the last refinement iteration for the first 3 splits on the FMNIST clustering tree already exhibited in Figure 5 in the main paper.

\section{Implementation Details}
\label{section:implementation_details}

This section describes our experiment settings as well as configurations necessary to reproduce our results.

\subsection{Deep Learning Framework and Hardware}

We have used PyTorch \cite{pytorch}, a well known framework for developing deep learning models. The hardware settings of relevance for this work are the following: Nvidia RTX 3070 GPU, Intel Core i5-7600K CPU.

\subsection{Code}

The code for our experiments was provided as a .zip file alongside this supplementary material. A README file in the code directory specifies how to use the code and the main Python packages necessary to run it.

\subsection{Reporting Performance}

For the main results reported (Table 1 in the paper), we selected the best of 5 tries, with slight modifications in hyperparameters in each try.

\subsection{Architecture}

\begin{table*}[t]
\footnotesize
\begin{center}
\begin{tabular}{rccccc}
\hline

Operation & Kernel & Strides & Feature Maps & LN? & Activation \\ \hline
Generator: $G(\mathbf{z})$ : $\mathbf{z} \sim \mathrm{Uniform}[0,1]$ &  &  & 100 &  \\
Fully Connected & & & $\frac{img\_dim^2}{4^2}\times128$ & No & ReLU\\ 
Transposed Convolution & $4\times4$  & $2\times2$ & 64   & No  & ReLU \\ 
Transposed Convolution & $4\times4$  & $2\times2$ & 1   & No  & Tanh  \\ 

\hline

Discriminator: $D(\mathbf{x})$ & & &  &  \\
Convolution (Shared with $C$) & $5\times5$ & $2\times2$ & $128$ & Yes & Leaky ReLU \\ 
Convolution (Shared with $C$) & $5\times5$ & $2\times2$ & $256$ & Yes & Leaky ReLU \\ 
Convolution (Shared with $C$) & $5\times5$ & $2\times2$ & $512$ & Yes & Leaky ReLU \\ 
Fully Connected &  &  & 1 & No & Sigmoid \\ 

\hline

Classifier: $C(\mathbf{x})$ & & &  &  \\
Convolution (Shared with $D$) & $5\times5$ & $2\times2$ & $128$ & Yes & Leaky ReLU \\ 
Convolution (Shared with $D$) & $5\times5$ & $2\times2$ & $256$ & Yes & Leaky ReLU \\ 
Convolution (Shared with $D$) & $5\times5$ & $2\times2$ & $512$ & Yes & Leaky ReLU \\ 
Fully Connected &  &  & 2 & No & Softmax \\

\end{tabular}
\caption{Architecture settings.}
\label{tab:architecture}
\end{center}
\end{table*}

We used the same architecture for all datasets, shown in Table~\ref{tab:architecture}. $G$, $D$ and $C$ respectively refer to each generator, discriminator and classifier created, either for a raw split sub-block or for a refinement sub-block. LN is short for Layer Normalization \cite{layer_normalization}, a well known deep learning normalization technique that is also used for stabilizing GAN training \cite{layer_normalization_gans}.

It is important to notice that the convolution layers for $D$ and $C$ share the same weights, \textit{i.e.}, the same $\mathbf{x}_{feature} = \mathrm{conv3}(\mathrm{conv2}(\mathrm{conv1}(\mathbf{x})))$ will be received as input by the non-shared Fully Connected layers of $D$ and $C$. These convolution weights are shared in order to force the classifier to differentiate between examples in a higher-level feature space, which will be learned by $D$ in its adversarial game with $G$. Learning to perform the classification in a higher-level feature space is more desirable than doing it in the plain data space (for images, the pixel space), since it is less likely to lead to an overfitted classification that takes into account low-level information to distinguish between the generated samples, which in turn could degrade the quality of the clustering created by the classifiers inference on the real data afterwards. For the Stanford Online Products dataset, we only made a slight architectural change for the refinements, in which we also shared all convolution layers parameters across both  refinement groups (\textit{i.e}., for $D_l, C_l, D_m, C_m$), which helped in stabilizing the image generation. A more specific detail regarding these shared convolutions is that only the gradients of the discriminator's loss are used to update the weights for these layers, with the classifier's gradients not being necessary for this update. This is because we empirically verified that the discriminator's learning was enough to obtain a sufficiently separable feature space for $C$'s fully connected layer to perform its classification with high accuracy.

\subsection{Hyperparameters}

\begin{table*}[t]
\footnotesize
\begin{center}
\begin{tabular}{rccc}
\hline
Parameters Settings & MNIST & Fashion MNIST & Stanford Online Products\\
\hline

Batch size for real data  & 100  & 100  & 100    \\
Batch size for each generator & 100  & 100  & 100  \\
Number of epochs & 120 & 120 & 100  \\
Number of refinements per node & 6 & 8 & 4  \\
Slope of Leaky ReLU & 0.2  & 0.2  & 0.2  \\
Learning rate for generator & 0.0002  & 0.0002  & 0.0002   \\
Learning rate for discriminator & 0.0001  & 0.0001  & 0.0002  \\
Learning rate for classifier & 0.00002  & 0.00002  & 0.00002   \\
Adam Optimizer &  $\beta_1 = 0.5, \beta_2 = 0.999$  &  $\beta_1 = 0.5, \beta_2 = 0.999$  &  $\beta_1 = 0.5, \beta_2 = 0.999$   \\
Diversity parameter $\gamma$ & $1.0$  & $1.0$  & $1.0$   \\
Initial noise variance  & $1.0$  & $1.5$  & $1.0$   \\

\end{tabular}
\caption{Specific hyperparameter settings used for each dataset.}
\label{tab:specific_configs}
\end{center}
\end{table*}

Table \ref{tab:specific_configs} describes the main hyperparameters used for the clustering of each dataset. Most of these settings are well known configurations in deep learning tasks, and they were all chosen based on similar GAN architectures employed for other tasks on these datasets, with some slight fine-tuning modifications that provided a better stabilization for each GAN training along with a high enough accuracy result obtained by the classifiers while distinguishing between the generators (in the 90 \% - 100 \% range). The number of training epochs are the same for both the raw split training and for each refinement training. We set the number of minibatch updates for each epoch to $\sum_{i}^{N}s_{k,i}$, for a GAN or MGAN receiving its real image samples according to $\mathbf{s}_{k}$. Unusual configurations worth explaining are: the the diversity parameter $\gamma$ and the initial noise variance. \textbf{Diversity parameter $\gamma$}: regularization parameter used on the generator for weighting the contribution of the classification loss vs the contribution of the adversarial loss, which is shown in line 11 of Algorithm 2 and line 8 of Algorithm 4 in the main paper. \textbf{Initial noise variance}: Adding Gaussian noise (with variance that linearly decays during the training epochs) to both generated and real images is a known stabilization practice for GAN training, as explained in \cite{gan_noise}, \cite{instance_noise_gans}.

\section{Drawbacks of the Method}

In this section, we discuss the results in more detail pointing to aspects that may lead to further improvements.

\subsection{Running Time}

The main drawback of our model consists in the complexity of sequentially training multiple different GAN/MGAN modules, especially in regard to running time. Each split in the tree creates one MGAN module for the raw split, and two new GAN modules for each refinement iteration. We performed at most 8 refinement iterations for our experiments with FMNIST, meaning that 16 GANs are trained for the refinements happening in each split. For creating a tree that reaches 10 leaf nodes, which was the case for our experiments, we need 9 splits. So, in total, $9 \times 1$ MGAN modules and $12 \times 9$ GAN modules were created for the FMNIST experiment, each one being trained to learn to represent from scratch its input distribution given by $\mathbf{s}_k$. It should be noted nonetheless that we set the number of minibatch samples for each epoch to $\sum_{i}^{N}s_{k,i}$, for a GAN or MGAN receiving its real image samples according to $\mathbf{s}_{k}$, and because the probability mass in $\mathbf{s}_k$ decreases at each newly created node, the number of minibatches will also decrease and hence the training time per epoch will also decrease for a GAN/MGAN at each new node. Even then, training so many GANs/MGANs still takes a large amount of time.

We have tried to overcome the running time issue with the following method: for a refinement iteration of a certain $\mathbf{s}_k$, instead of creating new GANs at each iteration $t$ and training the components from scratch, we preserve the weights learned by the GAN at the last iteration $t$. The reasoning behind this is that the GAN at refinement $t$ can be trained for less epochs, since it already starts its training with a previously learned representation of the data, having only to improve on it based on the newly estimated $\mathbf{s}_k^{(t)}$. This worked well for some nodes of the tree, yielding the same clustering result we achieved before, but with considerably less training epochs per refinement. But for some other nodes, after a certain number of refinements, the Generators' loss started to increase excessively, creating instability that compromised the image generation and thus worsened the clustering result. We still do not fully understand why this happens, but we believe that this method might work better with some other type of architecture design, especially one that employs a loss function with better robustness and convergence guarantees, such as the Wasserstein loss, instead of the common non-saturating loss used in our work.

\subsection{Hyperparameter and Architecture Choices}

The excessive time required to train so many GAN/MGAN modules discussed above constitutes in itself a limitation for the search of a set of unique hyperparameters and architectures that work well for each GAN/MGAN. But other than that, each GAN/MGAN is trained with different data distributions for each node of the tree, with each distribution becoming increasingly more homogeneous and thus simpler as nodes are created further away from the root. This difference in distribution complexity between different nodes means that the same set of hyperparameters might not be ideal to train all GANs/MGANs. It would be desirable that the hyperparameter tuning for each GAN/MGAN reflected the complexity of its respective distribution, \textit{e.g.}, for a simpler and more uniform data distribution, we could decrease the capacity of the GAN/MGAN architecture receiving it, or perhaps increase the diversity parameter $\gamma$ to encourage the 2 generated distributions at each split to become more distinct from each other. 

\end{document}